\documentclass[10pt, conference, letterpaper]{IEEEtran}
\IEEEoverridecommandlockouts
\usepackage{cite}
\usepackage{amsmath,amssymb,amsfonts}
\usepackage{algorithmic}
\usepackage{graphicx}
\usepackage{textcomp}
\usepackage{xcolor}

\usepackage{multirow}
\usepackage{makecell}
\usepackage{booktabs}

\usepackage[left=0.673in,right=0.673in,top=0.751in,bottom=1.10in]{geometry}

\def\BibTeX{{\rm B\kern-.05em{\sc i\kern-.025em b}\kern-.08em
    T\kern-.1667em\lower.7ex\hbox{E}\kern-.125emX}}
\begin{document}

% \title{ Paper Title*\\
\title{ Few-shot Message-Enhanced Contrastive Learning for Graph Anomaly Detection\\
}

\author{\IEEEauthorblockN{1\textsuperscript{st} Fan Xu}
\IEEEauthorblockA{\textit{School of Computer Science and Technology} \\
\textit{University of Science and Technology of China}\\
Anhui, China \\
markxu@mail.ustc.edu.cn}
\and
\IEEEauthorblockN{2\textsuperscript{nd} Nan Wang*}
\IEEEauthorblockA{\textit{School of Software Engineering} \\
\textit{Beijing Jiaotong University}\\
Beijing, China \\
wangnanbjtu@bjtu.edu.cn}
\and
\IEEEauthorblockN{3\textsuperscript{rd} Xuezhi Wen}
\IEEEauthorblockA{\textit{School of Software Engineering} \\
\textit{Beijing Jiaotong University}\\
Beijing, China \\
22126399@bjtu.edu.cn}
\and
\IEEEauthorblockN{4\textsuperscript{th} Meiqi Gao}
\IEEEauthorblockA{\textit{School of Software Engineering} \\
\textit{Beijing Jiaotong University}\\
Beijing, China \\
23126427@bjtu.edu.cn}
\and
\IEEEauthorblockN{5\textsuperscript{th} Chaoqun Guo}
\IEEEauthorblockA{\textit{School of Software Engineering} \\
\textit{Beijing Jiaotong University}\\
Beijing, China \\
chqguo@bjtu.edu.cn}
\and
\IEEEauthorblockN{6\textsuperscript{th} Xibin Zhao*}
\IEEEauthorblockA{\textit{School of Software Engineering} \\
\textit{Tsinghua University}\\
Beijing, China \\
zxb@tsinghua.edu.cn}

% \thanks{$^{*}$Corresponding author}%
\thanks{$^{*}$Nan Wang and Xibin Zhao are the corresponding authors}%

}

\maketitle

\begin{abstract}

Graph anomaly detection plays a crucial role in identifying exceptional instances in graph data that deviate significantly from the majority. It has gained substantial attention in various domains of information security, including network intrusion, financial fraud, and malicious comments, et al. Existing methods are primarily developed in an unsupervised manner due to the challenge in obtaining labeled data. For lack of guidance from prior knowledge in unsupervised manner, the identified anomalies may prove to be data noise or individual data instances. In real-world scenarios, a limited batch of labeled anomalies can be captured, making it crucial to investigate the few-shot problem in graph anomaly detection. Taking advantage of this potential, we propose a novel few-shot Graph Anomaly Detection model called FMGAD (Few-shot Message-Enhanced Contrastive-based Graph Anomaly Detector). FMGAD leverages a self-supervised contrastive learning strategy within and across views to capture intrinsic and transferable structural representations. Furthermore, we propose the Deep-GNN message-enhanced reconstruction module, which extensively exploits the few-shot label information and enables long-range propagation to disseminate supervision signals to deeper unlabeled nodes. This module in turn assists in the training of self-supervised contrastive learning. Comprehensive experimental results on six real-world datasets demonstrate that FMGAD can achieve better performance than other state-of-the-art methods, regardless of artificially injected anomalies or domain-organic anomalies.

% These limited labeled anomalies serve as valuable priors and can significantly inform the model, which is often overlooked by existing methods. 

\end{abstract}


\section{Introduction}
% Graphs, as a form of structured data, are widely used to represent complex dependencies among instances in various domains such as social networks, finance, biology, and transportation, enabling systematic modeling. With the rapid development of industrial and internet technologies, the occurrence of anomalous instances has become increasingly frequent, including fraud in social networks and leakage of sensitive corporate information. As a result, the problem of graph anomaly detection has gained significant attention in both industrial and academic communities.

Graph serves as a versatile representation of structured data, facilitating systematic modeling of complex dependencies among instances. It has been widely used in diverse domains like social networks, finance, biology, and transportation\cite{b1,b2,b3}. The rapid progress of industrial and internet technologies has led to a surge in the frequency of anomalous instances, encompassing fraudulent activities within social networks and the unauthorized disclosure of sensitive corporate information. Consequently, graph anomaly detection has garnered substantial attention from both industrial and academic communities.

Graph neural networks (GNNs)\cite{b4} have made significant advancements in graph representation learning by extending deep learning methods to graph-structured data, and they have found wide applications in graph anomaly detection. Unlike traditional anomaly detection methods that focus on vector data, graph anomaly detection requires the simultaneous exploration of both node attribute information and graph structure information, which is challenging for conventional approaches\cite{b5}. While, leveraging GNNs for modeling complex graph-structured data allows for the joint encoding of intricate interactions among instances and their respective attribute features, thereby facilitating the identification of anomalous nodes. 

Due to the labor-intensive and time-consuming nature of acquiring labeled anomaly data, most existing models in graph anomaly detection are developed in an unsupervised manner. For instance, DOMINANT\cite{b6} proposed a deep autoencoder that utilizes graph convolutional networks (GCNs) to reconstruct attributes and structure, thereby enhancing detection performance. GAAN\cite{b7} employs generative adversarial networks and generates pseudo-anomalies by utilizing Gaussian noise for discriminative training. Furthermore, with the rise of self-supervised learning, graph anomaly detection methods based on contrastive learning have gained popularity. For example, CoLA\cite{b8} employs random walks for graph augmentation, constructs positive and negative pairs, and designs proxy tasks for contrastive learning. Research findings have demonstrated that contrastive learning-based graph anomaly detection methods have achieved state-of-the-art performance in unsupervised settings.

However, due to the complexity and diversity of anomalies, as well as the lack of guided supervision from prior knowledge, unsupervised methods may suffer from local optima or exhibit biased anomaly detection performance. Nowadays, domain experts have provided feedback indicating that obtaining a limited number of labeled anomalies is feasible\cite{b9}. These labeled anomalies can serve as prior knowledge to guide model training and have great potential for improving graph anomaly detection performance. However, detecting anomalies in a few-shot setting remains a significant challenge. Existing semi-supervised and positive-unlabeled (PU) learning methods\cite{b10} have not yielded satisfactory results in this task. They rely on a sufficient number of labeled anomaly samples, making it difficult to effectively utilize supervised information in few-shot scenarios. Recently, some methods utilize meta-learning\cite{b11} and cross-domain transfer learning approaches\cite{b12} to address the few-shot setting. For instance, GDN\cite{b13} incorporates a meta-learning algorithm across networks to transfer meta-knowledge from multiple auxiliary networks for few-shot network anomaly detection. However, these methods have requirements for auxiliary networks or datasets, which are often difficult to obtain in real-world scenarios.

To address the aforementioned challenges, we propose a Few-shot Message-enhanced Contrastive-based Graph Anomaly Detector (FMGAD) that combines the rational utilization of few-shot labels with self-supervised contrastive learning. FMGAD consists of two main modules: (i)Multi-view contrastive learning module adopts the core idea of multi-view contrastive learning to facilitate both intra-view and cross-view contrastive learning. (ii)Deep-GNN message-enhanced reconstruction module leverages spectral high-pass filtering to design a deep message-passing network, effectively utilizing the few-shot label information. This module assists the Multi-view Contrastive Learning Module in learning tailored representations for the anomaly detector. The framework of our approach is illustrated in Fig~\ref{framework}. To summarize, our main contributions are summarized as follows:

\begin{itemize}
    \item To ensure that the self-supervised module can learn an optimal representation, we employ graph augmentation to obtain multiple views, enabling contrastive learning within and across views.
    \item To effectively utilize the few-shot label information and leverage it to assist the training of contrastive learning, we propose a Deep-GNN Message-Enhanced Reconstruction Module that provides a sufficiently large receptive field for the few-shot labeled nodes.
    \item We conduct extensive experiments on six real-world datasets with synthetically injected anomalies and organic anomalies. The experimental results demonstrate the effectiveness of our approach in few-shot graph anomaly detection.
\end{itemize}

\section{Related Work}
In this section, we briefly describe the related work on (1) Graph Anomaly Detection; (2) Few-shot Graph Learning and (3) Graph Augmentation.

\subsection{Graph Anomaly Detection}
Like other graph-based methods, semi-supervised learning is the most common graph representation learning mode and is also used in the field of graph anomaly detection. SemiGNN\cite{b14} utilizes a hierarchical attention mechanism to better associate different neighbors and different views. BWGNN\cite{b15} designs a band-pass filter kernel function satisfying Hammond's Graph Wavelet, transmitting information in corresponding frequency bands separately. Since anomalies are difficult to obtain, most existing methods are based on unsupervised modes and are mainly divided into two types: graph autoencoder and self-supervised contrastive learning. GAE (Graph Autoencoder)\cite{b16} reconstructs node features using an Encoder-Decoder architecture and defines nodes with high reconstruction loss as anomalous. DOMINANT\cite{b6} simultaneously reconstructs both structural information, such as the adjacency matrix, and node attributes to calculate anomaly scores. In recent years, with the rise of self-supervised learning and proxy tasks, various contrastive learning strategies have been widely applied. CoLA\cite{b8} utilizes random walk sampling to perform graph augmentation and subsequently constructs positive and negative node-subgraph pairs for contrastive learning. GraphCAD\cite{b17} employs a global clustering algorithm to partition the entire graph into multiple parts, where nodes injected from other parts are regarded as pseudo-anomalies, forming negative pairs. GRADATE\cite{b18} adopts edge modification graph augmentation technique and incorporates three types of contrastive learning strategies: node-node, node-subgraph, and subgraph-subgraph. 

% Sub-CR (2022 IJCAI) utilizes Graph Diffusion and subgraph sampling for global and local graph augmentation, followed by intra-view and inter-view contrastive learning.

% Different from most existing researches, we study a novel problem of few-shot graph anomaly detection in this paper by leveraging limited labeled anomalies to enhance the self-supervised learning of other unlabeled data.

% In summary, the differences among self-supervised contrastive learning methods mainly lie in the adoption of different graph augmentation techniques, subgraph construction methods, and contrastive learning strategies.

\subsection{Few-shot Graph Learning}

In most real-world scenarios, only very limited labeled samples are often available due to expensive labeling costs. In view of this, graph few-shot learning and cross-network meta learning are proposed to solve the problem of performance degradation when facing limited labeled data to a certain extent. For instance, GDN\cite{b13} is equipped with a cross-network meta-learning algorithm that utilizes a small number of labeled anomalies to enhance statistically significant deviations between abnormal nodes and normal nodes on the network. Meta-PN\cite{b19} infers high-quality pseudo-labels on unlabeled nodes via a meta-learning label propagation strategy while achieving a large receptive field during training. However, cross-domain auxiliary datasets are not always available, thus many non-meta-learning strategies have been explored. ANEMONE-FS\cite{b20} contains two multi-scale comparison networks, where the consistencies between nodes and contextual embeddings are maximized for unlabeled node while minimized for labeled anomalies in a mini-batch. 

\begin{figure*}[!t]
\centering
\includegraphics[width=5.6in]{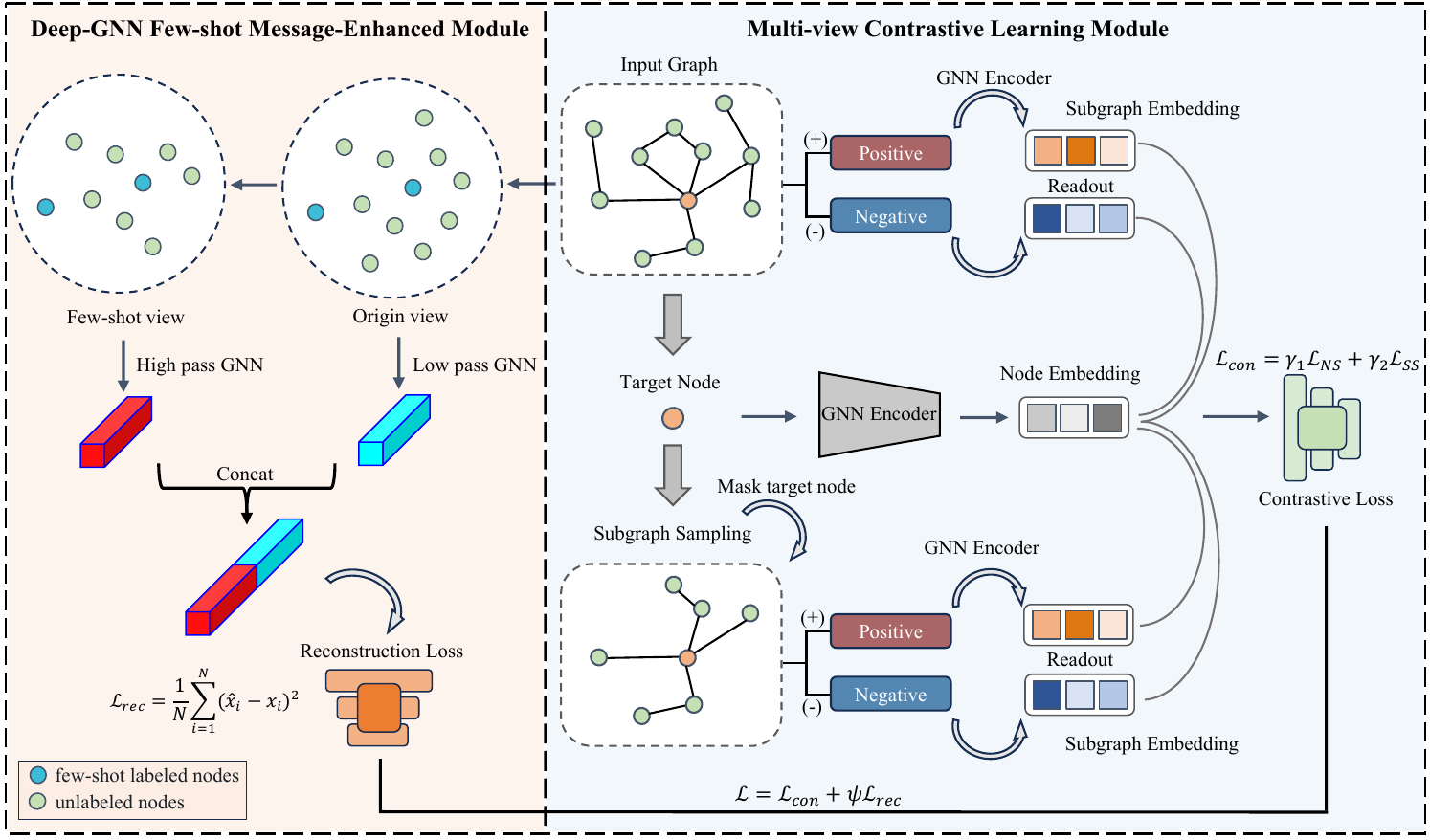}
\caption{The above image presents an overview of our model FMGAD, where the architecture demonstrates the details of the multi-view contrastive learning module(Right) and Deep-GNN few-shot message-enhanced module(Left) respectively.}
\label{framework}
\end{figure*}

\subsection{Graph Augmentation}
Similar to the vision domain, there are numerous augmentation methods in the field of graph representation learning. Specifically, graph augmentation techniques alter the attribute and structural characteristics of graph datasets within a certain range, providing convenience for self-supervised learning. The majority of existing methods focus on manipulating nodes or edges within the graph. These methods include: (i) enhancing by modifying or masking node features\cite{b21}, (ii) adapting the adjacency matrix or adjusting edge weights\cite{b22}, and (iii) utilizing Restarted Random Walk (RoSA)\cite{b23} to generate augmented local views.

\section{Problem Definition}
% $\mathcal{V} = \{v_{1},v_{2}, \dots, v_{n}\}$

In this section, we first introduce the notations mentioned in this paper, and then give the formal problem definition. Given an attributed graph $\mathcal{G} = (X,A)$, we denote its node attribute (i.e., feature) and adjacency matrices as $X\in\mathbb{R} ^{n\times d}$ and $\mathrm{A}\in\mathbb{R} ^{n\times n }$, where $\mathit{n}$ and $\mathit{d}$ are respectively the number of nodes and feature dimensions. It can also be defined as $\mathcal{G} = (V, E, X)$, where $\mathcal{V} = \{v_{1},v_{2}, \dots, v_{n}\}$ and $\mathcal{E} = \{e_{1},e_{2}, \dots, e_{n}\}$ represent node and edge sets respectively.

The definition of Few-shot GAD is to use the attribute and structure information of the graph to detect anomalies when a few-shot abnormal labeled nodes are known. We have a small set of labeled anomalies $\mathcal{V}^L$ and the rest set of unlabeled nodes $\mathcal{V}^U$ , where $|\mathcal{V}^L| << |\mathcal{V}^U|$, since the labeled anomaly nodes are difficult to obtain and few of them can be actually used. Then, our goal is to learn a model $\mathcal{F}(\cdot):\mathbb{R}^{N\times D}\to\mathbb{R}^{N\times1}$ on $\mathcal{V}^L\cup\mathcal{V}^U$, which measures node abnormalities by calculating their anomaly scores y.

% \begin{table}[h]     % t: 顶部 h: 当前位置
%   \caption{Notation summary}
%   \centering
%   \label{data}
%   \footnotesize
%   \begin{tabular}{cc}
%     \toprule
%     Notations & Definitions \\
%     \midrule
%     $\mathcal{G}$ & An undirected attributed graph  \\
%     $v_{i}$ & The $i$-th node of $\mathcal{G}$ \\
%     $\mathrm{A}\in\mathbb{R} ^{n\times n }$ & The adjacency matrix of $\mathcal{G}$ \\
%     $\mathrm{D}\in\mathbb{R} ^{n\times n }$ & The degree matrix of $\mathrm{A}$\\
%     $\mathrm{X}\in\mathbb{R} ^{n\times d }$ &  The feature matrix of $\mathcal{G}$\\
%     $\mathrm{H}^{\ell}\in\mathbb{R} ^{n\times d{}'}$ & The $\ell$-th layer hidden representation matrix\\
%     $h_{i}^{\ell}\in\mathbb{R} ^{1\times d{}'}$ & The $\ell$-th layer hidden representation of $v_{i}$\\
%     $\mathrm{W}^{\ell}\in\mathbb{R} ^{d{}'\times d{}'}$ & The $\ell$-th layer network parameters\\
%     % $\mathit{R}$ & The number of anomaly detection rounds\\
%     % $\mathit{S_{i} }$ & The final anomaly score of $v_{i}$\\
    
%   \bottomrule
% \end{tabular}
% \end{table}

\addtolength{\topmargin}{0.04in}

\section{Methodology}
In this section, we present the details of our proposed approach FMGAD for detecting graph node anomalies in few-shot scenarios. As shown in Fig~\ref{framework}, our approach mainly consists of two modules, including multi-view contrastive learning module and Deep-GNN message-enhanced reconstruction module. Graph anomalies are typically categorized as attribute-context anomalies and structural anomalies, and our method addresses both aspects. Firstly, we employ suitable graph augmentation techniques to construct different views and perform subgraph sampling for each target node. Next, to fully explore structural anomalies, we utilize proxy tasks and design a multi-view contrastive learning framework. Subsequently, to investigate features at the attribute-context level and leverage existing few-shot labels, we build a deep information augmentation reconstruction module. In all, our model starts from the essence of graph anomalies, designs self-supervised learning objectives, and incorporates supervised constraints using few-shot labels. In the rest of this section, we demonstrate the details of the whole framework respectively.

\subsection{Graph Augmentation}
The self-supervised strategy based on contrastive learning enables not only differentiation learning within the same scale, such as "node vs. node," but also discrimination across different scales, such as "node vs. subgraph." As discussed in related work, to ensure that the self-supervised learning module can extract rich attribute and structural information, it is necessary to design augmentation strategies and proxy tasks tailored to the current task. For graph anomaly detection, according to [reference], anomalies in graph nodes often manifest as a mismatch with their surrounding environment.

For several popular graph augmentation strategies in current graph representation learning, such as node feature perturbation or masking and edge modification. Including Graph Diffusion, it essentially involves perturbing the adjacency matrix and modifying the target edges. We argue that these strategies are not suitable for graph anomaly detection because they may alter the underlying logic or semantic features of the data. This could particularly have negative effects on detecting naturally occurring anomalies rather than artificially injected anomalies. Hence, we utilize random walks with restart (RWR) to obtain augmented views. Specifically, for each selected target node, we sample subgraphs of fixed size $p$. Unlike standard random walks, RWR introduces a restart probability, where there is a certain probability of restarting from the initial node at each step. Therefore, using RWR to sample subgraphs does not introduce additional anomalies.

\columnsep 0.241 in

\subsection{Multi-view Contrastive Learning Module}
Furthermore, we constructed a multi-view contrastive learning module. This module utilizes GNN encoders and decoders to perform contrastive learning between the target node and two views, simultaneously learning discriminative attribute and structural topological information. It consists of two parts: Node-Subgraph and Subgraph-Subgraph, capturing features within each view and across different views respectively.

\vspace{0.1cm}
\noindent \textbf{Node-Subgraph Contrast.} In each view, a target node $v_i$ forms a positive pair with its located subgraph and forms a negative pair with a random subgraph where another node $v_j$ is located. We first adopt a GCN encoder that maps the features of nodes in the subgraph to the embedding space. The hidden-layer representation can be defined as:

\begin{equation}
    H_{\omega }^{\ell+1} = GNN(A_{\omega }, H_{\omega }^{\ell}) = \sigma ( D_{\omega }^{-\frac12} A_{\omega } D_{\omega }^{-\frac12} H_{\omega }^{\ell} W^{\ell}),
\end{equation}
% \begin{equation}
%     H_{\omega _i}^{(\ell+1)} = GNN(\widetilde{A}_{\omega _i}, H_{\omega _i}^{(\ell)}) = \sigma ( \widetilde{D}_{\omega _i}^{-\frac12}\widetilde{A}_{\omega _i}\widetilde{D}_{\omega _i}^{-\frac12}H_{\omega _i}^{(\ell)}W_{\omega _i}^{(\ell)}),
% \end{equation}

\noindent where $H_{\omega }^{\ell+1}$ and $H_{\omega }^{\ell}$ denote the $(\ell+1)$-th and $\ell$-th layer hidden representation in view $\omega $, $\widetilde{D}_{\omega }^{-\frac12}\widetilde{A}_{\omega }\widetilde{D}_{\omega }^{-\frac12}$ is the normalization of the adjacency matrix in view $\omega _i$ and $W^{\ell}$ is the network parameters. It is noteworthy that the networks operating under two views employ identical architecture and parameter sharing.
Then we take the average pooling function as the readout module to obtain the subgraph-level embedding vector $e_{\omega}$:
\begin{equation}
    e_{\omega}= READOUT(H_{\omega})=\frac{1}{K}\sum_{j=1}^K (H_{\omega})_K,
\end{equation}

\noindent where $K$ denotes the number of remaining nodes in the subgraph. Given that the target node is masked within the subgraph, we utilize the weight matrix of the GCN encoder to project the features onto a shared embedding space. Mathematically, this can be formulated as follows:

\begin{equation}
    h_{\omega}^{\ell+1} = \sigma (h_{\omega}^{\ell}W^{\ell} ).
\end{equation}

In each view, the anomalous degree of a target node depends on its similarity to the paired subgraph embedding. Therefore, we choose a Bilinear model to quantify the relationship:

\begin{equation}
    s_{\omega} = sigmoid ( e_{\omega} W_s \, h_{\omega}^T ),
\end{equation}

\noindent where $W_s$ is a learnable matrix. We employ the binary cross-entropy loss to measure the contrastive loss in a single view that can be demonstrated as:

\begin{equation}
    \mathcal{L}_{NS}^{\omega}=-\sum_{i=1}^N\left(y_i\log\left(s_{\omega i} \right)+(1-y_i)\log\left(1-s_{\omega i}\right)\right),
\end{equation}

\noindent where $y_i$ is equal to 1 when $s_{\omega i}$ denotes a positive pair, and is equal to 0 when $s_{\omega i}$ denotes a negative pair. The same operations and model architecture are used on the second view, and both views share model parameters. Thus the final node-subgraph contrast loss is:
\begin{equation}
    \mathcal{L}_{NS}=\alpha\mathcal{L}_{NS}^1+(1-\alpha)\mathcal{L}_{NS}^2,
\end{equation}

\noindent where $\alpha\in(0,1)$ is a trade-off parameter to balance the importance between two views.

\vspace{0.1cm}
\noindent \textbf{Subgraph-Subgraph Contrast.} Instead of intra-view contrast, subgraph-subgraph contrast implements cross-view contrastive learning. It aims to learn more representative subgraph embeddings, thereby enhancing the neighborhood representations of target nodes. Specifically, a subgraph establishes a positive pair with the subgraph formed by its target node $v_i$ in another view, while it forms negative pairs with two subgraphs where another node $v_j$ is located in both views. Inspired by [], we employ a loss function to optimize the contrast:

\begin{equation}
    \mathcal{L}_{SS}=-\sum_{i=1}^n\log\frac{\exp\left(\boldsymbol{e}_{1i}\cdot\boldsymbol{e}_{2i}\right)}{\exp\left(\boldsymbol{e}_{1i}\cdot\boldsymbol{e}_{1j}\right)+\exp\left(\boldsymbol{e}_{1i}\cdot\boldsymbol{e}_{2j}\right)},
\end{equation}

\noindent where $\boldsymbol{e}_{1i}$ and $\boldsymbol{e}_{1i}$ denote the embeddings of the subgraphs that the target node $v_i$ belongs to in two views, $\boldsymbol{e}_{1j}$ and $\boldsymbol{e}_{1j}$ represent the embeddings of the subgraphs of another node $v_j$ separately. Then the final multi-view contrastive loss is:

\begin{equation}
    \mathcal{L}_{con}=\gamma \mathcal{L}_{NS}+ (1-\gamma)\mathcal{L}_{SS},
\end{equation}

\noindent where $\gamma \in(0,1)$ balances the influence of two contrastive learning modes.

\subsection{Deep-GNN Message-Enhanced Reconstruction Module}
In the context of few-shot scenarios, the availability of anomaly label information is severely limited. Conventional semi-supervised graph anomaly detection methods suffer from the issue of over-smoothing, making it challenging to extend the receptive field and effectively propagate label information to deeper neighborhoods. To address this challenge, we propose leveraging the concept of AutoEncoder from unsupervised methods to reconstruct attributes. Additionally, we introduce a scalable deep graph neural network (GNN) architecture to enhance the utilization of few-shot labels and their associated features, thereby improving the performance of anomaly detection in graph data.

Initially, we extract a few-shot environmental subgraph from the original graph, comprising a subgraph originating from the few-shot labeled node and encompassing its M-order neighbors. To facilitate the sparse message enhanced feature reconstruction process, distinct graph neural network (GNN) architectures are employed for encoding the original graph and the few-shot environment subgraph. In particular, for the original view, GNN encoder is with low-pass filtering characteristics, such as GCN, GAT, GIN. These GNN models effectively capture and propagate information within the graph, enabling accurate attribute reconstruction and subsequent anomaly detection. The transform of corresponding GNN encoder is as follows:

\begin{equation}
    H_r^{\ell+1} = \sigma (D^{-1/2}AD^{-1/2} H_r^{\ell} W_r).
\end{equation}

To leverage the specific attributes of sparse anomaly samples within the few-shot environment subgraph and their high-order correlation with the surrounding context, we propose a scalable deep graph neural network (Deep-GNN)\cite{b24} architecture that enables long-range propagation. This approach allows for the consideration of a broader range of context nodes, thereby expanding the receptive field of sparse anomaly samples. To address the challenge of over-smoothing that arises when increasing the propagation step size in GNN, we introduce a high-pass filtering GNN\cite{b25} that operates in the spectral domain:

\begin{equation}
    \mathcal{F}_H=\varepsilon I-D^{-1/2}AD^{-1/2}=(\varepsilon-1)I+L,
\end{equation}

\begin{equation}
    H_f^{\ell+1} = \sigma (\mathcal{F}_H H_f^{\ell} W_f).
\end{equation}

According to\cite{b25}, high-pass filtering GNN can overcome the over-smoothing problem to a certain extent, and therefore can be extended to more layers. Then we concatenate the node embeddings obtained from the original graph and the few-shot environmental subgraph:

\begin{equation}
    H=CONCAT(H_r,H_f),
\end{equation}

\noindent for nodes that do not appear in the few-shot environment subgraph, their $h_{f}$ is padded with 0. Then a layer of MLP is applied to obtain the reconstructed node embeddings:

\begin{equation}
    \widehat{X}=MLP(H).
\end{equation}

The reconstruction loss of the original graph is calculated by MSE loss:

\begin{equation}
    \mathcal{L}_{rec}=\frac1N\sum_{i=1}^N(\widehat{x}_i - x_i)^2.
\end{equation}

% By incorporating the high-pass filtering mechanism, our GNN model effectively captures and emphasizes the high-frequency components of the graph, facilitating the preservation of local structural information and preventing the loss of crucial details during propagation. This design choice enhances the discriminative power of the GNN, enabling it to identify and distinguish sparse anomaly samples within the few-shot environment subgraph accurately. The incorporation of long-range propagation and high-pass filtering within our GNN framework contributes to a more robust and effective anomaly detection mechanism in graph data.

\subsection{Anomaly Detector}

To jointly train the multi-view contrastive learning module and the Deep-GNN message-enhanced reconstruction module, we optimize the following objective function:

\begin{equation}
    \mathcal{L} =  \mathcal{L}_{con} + \psi \mathcal{L}_{rec},
\end{equation}

\noindent where $\psi$ is a controlling parameter which balances the importance of the two modules. By minimizing the above objective function, we can compute the anomaly score of each node.

\section{Experiments}
In this section, we conduct empirical evaluations to showcase the efficacy of the proposed framework. Our primary objective is to address the following research inquiries:

\begin{itemize}
    \item \textbf{RQ1.} Can our method perform well in extreme few-shot scenarios?
    \item \textbf{RQ2.} How our model behave when changing the degree of label availability and the number of Deep-GNN layers?
    \item \textbf{RQ3.} How do the key designs and components influence the performance of our method?
\end{itemize}

\subsection{Experimental Settings}

\noindent \textbf{Dataset.} To thoroughly evaluate our method's performance in identifying both naturally occurring organic anomalies and artificially injected anomalies, we selected two categories of datasets. The first category consists of two authentic datasets: Cora\cite{b26} and Citeseer\cite{b27}, that do not inherently contain organic anomalies but require manual injection of anomalies. The second category comprises three authentic datasets: Wiki\cite{b28}, Reddit\cite{b29} and YelpChi\cite{b30}, that inherently contain organic anomalies. For anomaly injection, we followed the same approach as DOMINANT by injecting the same number of feature and structural anomalies into the three datasets that previously did not have any organic anomalies.

\begin{table}[h]     % t: 顶部 h: 当前位置
  \caption{Statistics of Datasets}
  \centering
  \label{data}
  \footnotesize
  \begin{tabular}{c|cc p{0.9cm}<{\centering} p{0.9cm}<{\centering} p{0.9cm}<{\centering}}
    \toprule
    Dataset & Nodes & Edges & Features & Anomaly & Ratio(\%)  \\
    \midrule
    Cora & 2,708 & 5,429 & 1433 & 150 & 5.54  \\ % 150
    CiteSeer & 3327 & 10,154 & 3703 & 150 & 4.51 \\ % 150
    % PubMed & 19,717 & 44,338 & 500 & 600 & 3.04  \\ % 600
    Wiki & 9,227 & 18,257 & 64 & 217 & 2.35  \\ % 217
    Reddit & 15,860 & 136,781 & 602 & 796 & 5.02  \\  % 796
    YelpChi & 23,831 & 98,630 & 32 & 1,217 & 5.11  \\
    
  \bottomrule
\end{tabular}
\end{table}

\begin{table*}[h]
  \centering
  \caption{Performance comparison results (10-shot) w.r.t. AUC-ROC and AUC-PR on five datasets.}
  \label{results}
  \begin{tabular}{p{1.9cm}<{\centering}|p{1.1cm}<{\centering} p{0.9cm}<{\centering}|p{1.1cm}<{\centering} p{0.9cm}<{\centering} | p{1.1cm}<{\centering} p{0.9cm}<{\centering} | p{1.1cm}<{\centering} p{0.9cm}<{\centering}|p{1.1cm}<{\centering} p{0.9cm}<{\centering}}
    \toprule
    \multicolumn{1}{c|}{\multirow{2}{*}{Methods}} & \multicolumn{2}{c|}{Cora}   & \multicolumn{2}{c|}{Citeseer} & \multicolumn{2}{c|}{Wiki} & \multicolumn{2}{c|}{Reddit} & \multicolumn{2}{c}{YelpChi}\\
    \cline{2-11}
    \multicolumn{1}{c|}{} & \scriptsize{AUC-ROC} & \scriptsize{AUC-PR} & \scriptsize{AUC-ROC} & \scriptsize{AUC-PR} & \scriptsize{AUC-ROC} & \scriptsize{AUC-PR} & \scriptsize{AUC-ROC} & \scriptsize{AUC-PR} & \scriptsize{AUC-ROC} & \scriptsize{AUC-PR} \\
    \hline
    GCN & 0.5239 & 0.0427 & 0.4128 & 0.055 & 0.4324 & 0.0239 & 0.4975 & 0.0826 & 0.3371 & 0.0725 \\
    GAT & 0.5473 & 0.0495 & 0.4645 & 0.062 & 0.4373 & 0.0284 & 0.5184 & 0.1225 & 0.3564 & 0.0834 \\ 
    SemiGNN & 0.6637 & 0.1293 & 0.5322 & 0.074 & 0.4785 & 0.0332 & 0.6249 & 0.1953 & 0.4146 & 0.1378 \\
    BWGNN & 0.6855 & 0.1876 & 0.5421 & 0.081 & 0.4668 & 0.0295 & 0.5863 & 0.1634 & 0.5473 & 0.2161\\
    \hline
    DOMINANT & 0.7483 & 0.2741 & 0.8279 & 0.2415 & 0.4488 & 0.0227 & 0.7429 & 0.3185 & 0.4872 & 0.1652\\
    CoLA & 0.7515 & 0.2398 & 0.8738 & 0.2942 & 0.5373 & 0.0319 & 0.7257 & 0.2474 & 0.3985 & 0.1579\\
    GraphCAD & 0.7674 & 0.2892 & 0.8521 & 0.2787 & 0.5282 & 0.0249 & 0.7536 & 0.2643 & 0.4238 & 0.1843\\
    GRADATE & 0.7786 & 0.2973 & 0.8872 & \underline{0.3471} & \underline{0.5471} & 0.0322 & 0.7472 & 0.2879 & 0.4994 & 0.2164 \\
    \hline   
    GDN & 0.7736 & 0.1965 & 0.7963 & 0.1826 & 0.5248 & 0.0326 & \underline{0.8136} & 0.3084 & 0.7281 & 0.2785 \\
    Meta-PN & 0.8537 & 0.2817 & 0.8127 & 0.2273 & 0.4663 & 0.0276 & 0.8064 & 0.3126 & 0.7549 & 0.2698 \\
    ANEMONE-FS & \underline{0.8836} & \underline{0.3062} & \underline{0.9028} & 0.3294 & 0.5317 & \underline{0.0348} & 0.8123 & \underline{0.3352} & \underline{0.7729} & \underline{0.2977} \\
    \hline
    FMGAD & \textbf{0.8928} & \textbf{0.3187} & \textbf{0.9193} & \textbf{0.3981} & \textbf{0.6133} & \textbf{0.0438} & \textbf{0.8326} &\textbf{0.3561} & \textbf{0.8052} & \textbf{0.3338} \\
  \bottomrule
\end{tabular}
\end{table*}

\noindent \textbf{Compared Methods.} We compare our proposed method FMGAD with other three categories of methods. (i) Conventional semi-supervised GNN models: GCN\cite{b4}, GAT\cite{b31}, and semi-supervised methods designed for GAD: SemiGNN\cite{b14}, BWGNN\cite{b15}. (ii) Unsupervised GNN-based graph anomaly detection methods: DOMINANT\cite{b6}, CoLA\cite{b8}, GraphCAD\cite{b17} and GRADATE\cite{b18}. (iii) Few-shot methods on graph anomaly detection: GDN\cite{b13}, Meta-PN\cite{b19} and ANEMONE-FS\cite{b20}.

\noindent \textbf{Evaluation Metrics.} We employ two popular and effective metrics for evaluation, the Area Under Receiver Operating Characteristic Curve (AUC-ROC) and the Area Under Precision-Recall Curve (AUC-PR)\cite{b32}. AUC-ROC quantifies the ability of a binary classifier by measuring the area under the receiver operating characteristic curve. AUC-PR captures the trade-off between the two metrics and is particularly useful when the dataset is imbalanced or when the focus is on positive instances.

\noindent \textbf{Implementation Details.}
% All our experiments are conducted with a 24 GB 3090 GPU. The proposed FMGAD is implemented in PyTorch. We use a 1-layer MLP with 64 hidden units for the feature-label transformer. We apply $L_2$ regularization with $\lambda $ = 0.01 on the weights of the first neural layer and set the dropout rate for both neural layers to be 0.3. For all datasets, we set the number of few-shot labeled anomalies as 10.

All our experiments are conducted with a 24 GB 3090 GPU, and the proposed FMGAD is mainly implemented through pyg library. In our implementation, the size K of subgraph of each target node and the dimension of hidden layer are fixed to 8 and 128, respectively. In the contrastive learning module, the GNN network is set to 2 layers; in the reconstruction module, the low-pass and high-pass GNN Encoder are set to 2 and 5 layers. For each dataset, we set the number of few-shot labeled anomalies as 10, and the trade-off parameters $\alpha, \gamma_1, \gamma_2, \psi$ are chosen as 0.7, 0.6, 0.4, and 0.5 separately. 

% The model is trained using the Adam optimizer with a learning rate of 0.001.

\subsection{Experimental Results (RQ1)}

% In this subsection, we evaluate our proposed model FMGAD and the remaining baseline methods in terms of AUC-ROC and AUC-PR. Table 3 shows the performance comparison of each method on the artificially injected anomaly and organic anomaly datasets. Based on the experimental results, we have the following observations:

In this subsection, we consider semi-supervised, unsupervised and other few-shot baseline methods for comparing with our methods in terms of AUC-ROC and AUC-PR. To ensure few-shot scenarios, for all few-shot GAD methods, we use 10 annotated anomalies during model training. Tab~\ref{results} shows the overall performance comparison on both artificially injected anomaly and organic anomaly datasets. FMGAD consistently outperforms all baseline methods on all six real-world datasets, thereby validating the effectiveness of our approach in addressing anomaly detection in few-shot scenarios. Based on the experimental results, we have the following observations:

\begin{itemize}
    \item Conventional semi-supervised graph anomaly detection methods (i.e., GCN, GAT, SemiGNN, and BWGNN) generally do not exhibit competitive performance, indicating their limited ability to exploit label information. It performs even worse than unsupervised methods on almost all datasets. This discrepancy can be attributed to the reliance of conventional semi-supervised methods on sufficient label information for message propagation, which exacerbates the over-smoothing issue in few-shot scenarios and hinders the learning of abnormal features. However, unsupervised methods leverage AutoEncoder or contrastive learning strategies to uncover deep data distributions based on local features and structures. Thus, they can achieve strong discrimination capabilities when it comes to identifying artificially injected anomalies.
    \item On datasets with artificially injected anomalies, the unsupervised methods achieve performance that matches existing few-shot graph anomaly detection methods. However, on organic anomaly datasets, unsupervised methods generally underperformed compared to few-shot methods. In particular, compared to the GRADATE, on YelpChi dataset, our FMGAD has 60.35\% and 54.25\% improvement w.r.t. AUC-ROC and AUC-PR, respectively. This is most likely because real data often possesses numerous expert priors, and unsupervised methods tend to blindly map and partition features.
    \item In comparison to existing few-shot graph anomaly detection methods, our approach has demonstrated notable advancements. To be specific, on Wiki dataset, our method FMGAD outperforms GDN by 16.86\% and 34.36\% in terms of AUC-ROC and AUC-PR, respectively. The three methods we compared are all founded on meta-learning principles, and the efficacy of meta-learning methods relies heavily on the quality of the auxiliary network or dataset. However, in many real-world scenarios, datasets often do not meet such stringent requirements.
\end{itemize}

% 对比现有的few-shot图异常检测方法，我们的方法也取得了一定程度上的提升。所对比的三个方法都是基于元学习开发的，元学习方法性能很大程度上依赖辅助网络或数据集的质量。而对于大多数现实场景，数据集一般是不具备这样的需求的。使用已知的few-shot异常节点作为元知识

\subsection{Sensitivity \& Robustness Analysis (RQ2)}

% We evaluate FMGAD in few-shot learning scenario (k = 10) where 10 annotated anomalies are known during model training.
In order to verify the effectiveness of FMGAD in different few-shot anomaly detection settings, we change the number $k$ of anomalous samples for model training to form k-shot learning settings for evaluation. Specifically, we perform experiments on all five datasets and select $k$ from \{1, 3, 5, 10, 15, 20\}. The experimental results are summarized in Tab~\ref{few}.

\begin{table}[h]
\centering
% \small
  \caption{Few-shot Performance analysis of FMGAD.}
  \label{few}
  \begin{tabular}{p{1.0cm}<{\centering} | p{1.0cm}<{\centering} p{1.0cm}<{\centering} p{0.8cm}<{\centering} p{1.0cm}<{\centering} p{1.0cm}<{\centering}}
    \toprule
    Setting & Cora & Citeseer & Wiki & Reddit & YelpChi \\
    % &  F1-macro & AUC & F1-macro & AUC \\
    \midrule
    1-shot & 0.8681 & 0.9039 & 0.5854 & 0.8216 & 0.7667 \\
    3-shot & 0.8843 & 0.9126 & 0.6003 & 0.8244 & 0.7786 \\
    5-shot & 0.8906 & 0.9177 & 0.6078 & 0.8263 & 0.7928 \\
    10-shot & \textbf{0.8946} & \underline{0.9193} & \underline{0.6133} & \underline{0.8326} & \underline{0.8052} \\
    15-shot & \underline{0.8921} & \textbf{0.9225} & \textbf{0.6158} & \textbf{0.8367} & \textbf{0.8127} \\
    % 20-shot & 0.8956 & 0.9237 & 0.6179 & 0.8392 & 0.8093 \\
  \bottomrule
\end{tabular}
\end{table}

As observed, even in scenarios where only 1-shot anomalies are provided, FMGAD can still outperform other baseline methods, demonstrating its superior performance. For instance, on Reddit dataset, the FMGAD with 1-shot anomaly outperforms GraphCAD by 9.02\% in terms of AUC-ROC. When compared with ANEMONE-FS, it achieves improvements of 6.52\% in terms of AUC-ROC with 5-shot anomalies. This demonstrates the effectiveness of the FMGAD method for extremely limited anomalous labels. Furthermore, we also observe that as the number of few-shot anomaly labels increases, FMGAD's performance generally improves, which further confirms the effectiveness of our method.

\begin{figure}[h]
\centering
\includegraphics[width=0.48\textwidth]{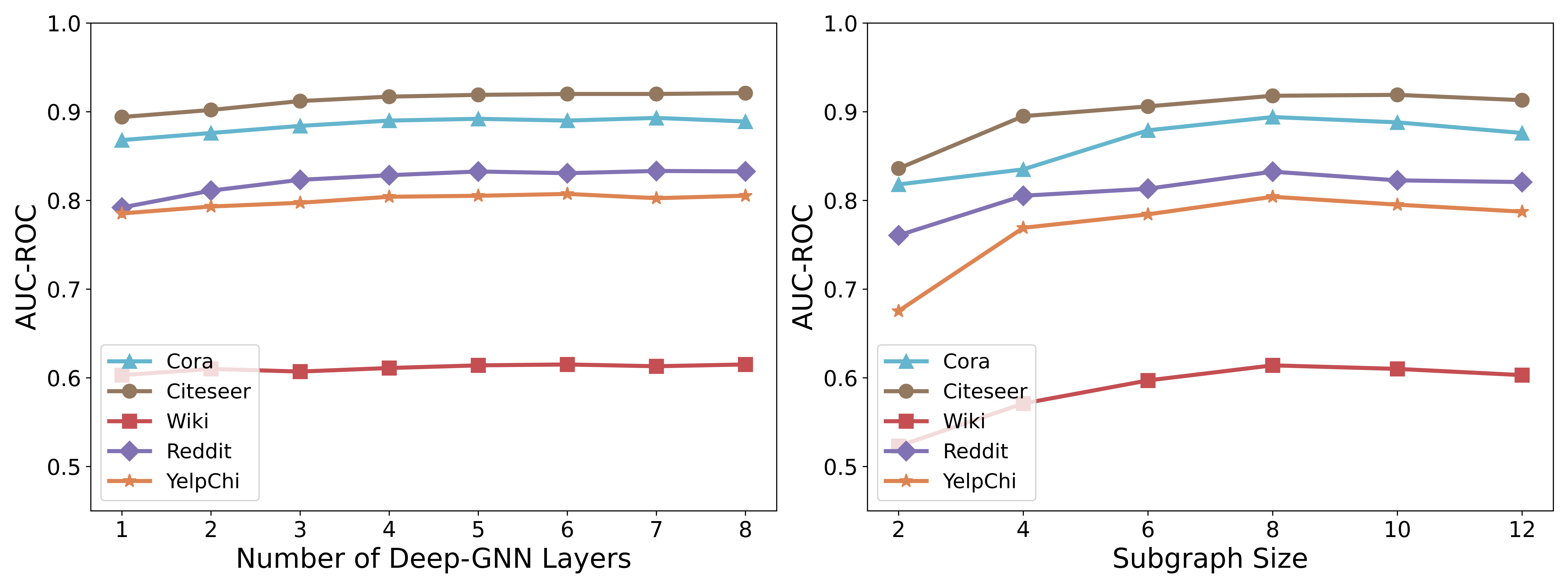}
\caption{Performance with different number of Deep-GNN layers and the size of subgraph sampled by RWR.}
\label{order}
\end{figure}

Subsequently, we investigated the effects of varying the number of Deep-GNN layers in the reconstruction module and adjusting the number of nodes through RWR sampling in the enhanced subgraph sampling of the contrastive learning module on model performance. The corresponding experimental results are shown in Fig~\ref{order}.

% 观察左图发现，随着采样子图节点数量的增加，模型性能先是上升，在一定程度后下降。这是因为，当目标节点采样子图太小时，模型难以捕捉数据的局部结构特征，导致性能较差。而当采样子图太大时，子图将包含冗余信息，这将损害模型性能。
Analyzing the image on the left, we observe a trend where the model performance initially improves with an increasing number of sampling subgraph nodes. However, beyond a certain threshold, further increments in the number of nodes lead to a diminishing effect on the model's performance. This is because insufficient sampling of the target node subgraph makes it challenging for the model to capture the local structural characteristics of the data, leading to subpar performance. Conversely, if the sampled subgraph is excessively large, it may contain redundant information, thereby adversely affecting model performance. Observing the graph on the right, we note that with an increase in the number of Deep-GNN layers, the model performance exhibits a slight improvement initially, followed by a subsequent decline. We attribute the performance improvement to the Deep-GNN network effectively propagating label information to more distant neighbors within the graph. However, an excessive number of layers will inevitably introduce the challenge of over-smoothing, which can negatively impact the model's performance. Hence, finding an optimal balance in the size of the sampled subgraph and the number of Deep-GNN layers is crucial for achieving optimal results.

\subsection{Ablation Study (RQ3)}

\begin{figure}[h]
\centering
\includegraphics[width=0.47\textwidth]{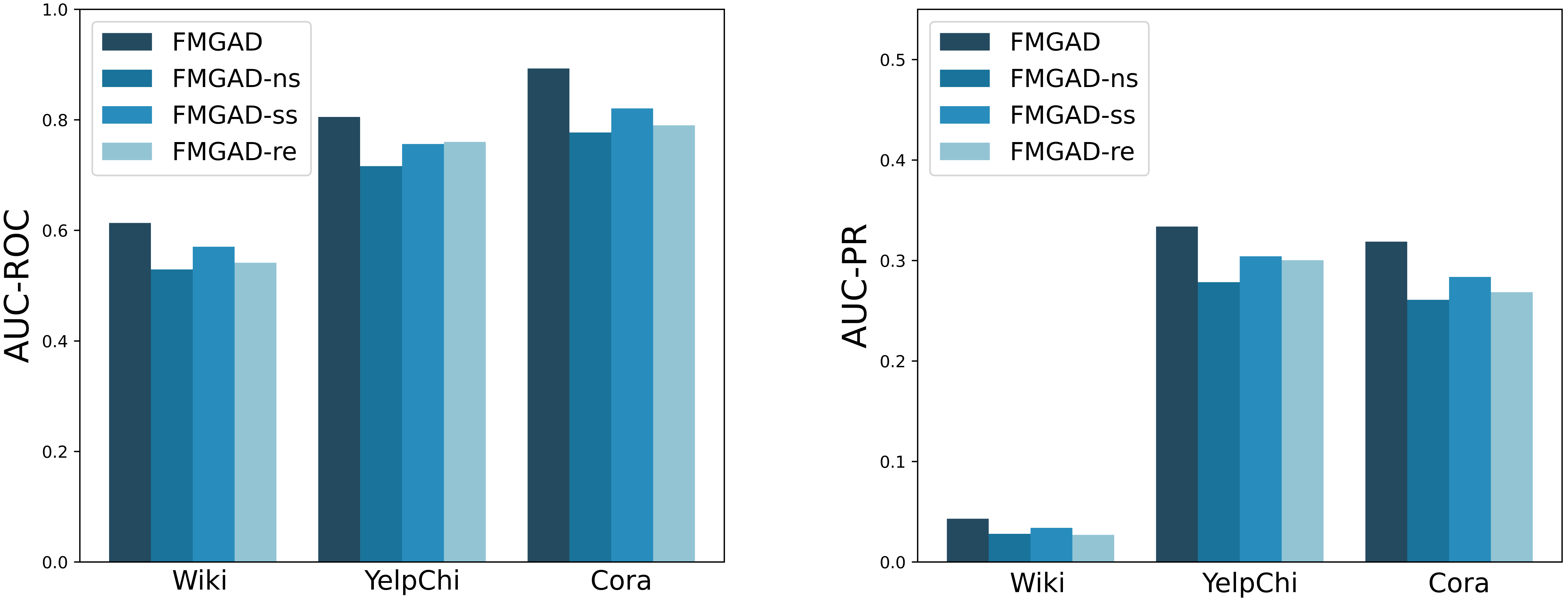}
\caption{Ablation Performance on different variants.}
\label{ablation}
\end{figure}

In order to verify the effectiveness of each key component of FMGAD, we conduct an ablation study on the variants of the proposed approach. Concretely, we introduce three variants of our approach: FMGAD-ns and FMGAD-ss, which individually exclude the node-subgraph and subgraph-subgraph contrastive learning sub-modules, and FMGAD-re, which omits the Deep-GNN few-shot message-enhanced module. The detailed results are shown in Fig~\ref{ablation}.

As observed above, for each variant that excludes a specific module, there has been a noticeable degradation in the model's performance. Among these variants, FMGAD-ns stands out as the most significantly impacted, as it eliminates the node-subgraph contrastive sub-module. Specifically, it drops by 8.62\% and 13.17\% on YelpCHi datasets in terms of AUC-ROC and AUC-PR. In summary, through ablation studies, we affirm the robustness and efficacy of our proposed technique in addressing graph anomaly detection under few-shot scenarios.

\section{Conclusion}

In this paper, we investigate the problem of graph anomaly detection in few-shot scenarios. Through a comprehensive analysis of existing semi-supervised, unsupervised, and customized few-shot methods, we propose FMGAD, a novel anomaly detector that combines few-shot message enhancement with multi-view self-supervised contrastive learning. Our model effectively utilizes the self-supervised contrastive learning strategy to capture local structures and features within the graph. Additionally, we introduce a deep message-passing mechanism that incorporates high-pass convolutional filtering functions to enable deep propagation of few-shot node information. Extensive experiments conducted on multiple real-world datasets demonstrate the outstanding performance of FMGAD.

\section*{Acknowledgment}

This work was supported in part by the Supported by the Fundamental Research Funds for the Central Universities under Grant 2022RC026; in part by the CCF-Tecent Open Fund under Grant CCF-Tencent RAGR20220112; in part by the CCF-NSFOCUS Open Fund; in part by the NSFC Program under Grant 62202042, Grant 62076146, Grant 62021002, Grant U20A6003, Grant U19A2062, Grant 62127803, and Grant U1911401.


\begin{thebibliography}{00}
\bibitem{b1} Ma X, Wu J, Xue S. A comprehensive survey on graph anomaly detection with deep learning[J]. IEEE TKDE, 2021.
\bibitem{b2} Ding K, Shu K, Shan X. Cross-domain graph anomaly detection[J]. IEEE TNNLS, 2021, 33(6): 2406-2415.
\bibitem{b3} I. S. Jacobs and C. P. Bean, ``Fine particles, thin films and exchange anisotropy,'' in Magnetism, vol. III, G. T. Rado and H. Suhl, Eds. New York: Academic, 1963, pp. 271--350.
\bibitem{b4} Kipf T N, Welling M. Semi-supervised classification with graph convolutional networks[J]. ICLR, 2016.
\bibitem{b5} Chandola V, Banerjee A, Kumar V. Anomaly detection: A survey[J]. ACM computing surveys (CSUR), 2009, 41(3): 1-58.
\bibitem{b6} Kaize Ding, Jundong Li, Rohit Bhanushali, and Huan Liu. 2019. Deep anomaly detection on attributed networks. In Proceedings of the 2019 SIAM International Conference on Data Mining. SIAM, 594–602.
\bibitem{b7} Zhenxing Chen, Bo Liu, Meiqing Wang, Peng Dai, Jun Lv, and Liefeng Bo. 2020. Generative adversarial attributed network anomaly detection. In Proceedings of the 29th ACM International Conference on Information \& Knowledge Management. 1989–1992.
\bibitem{b8} Yixin Liu, Zhao Li, Shirui Pan, Chen Gong, Chuan Zhou, and George Karypis. 2021. Anomaly detection on attributed networks via contrastive self-supervised learning. IEEE TNNLS 33, 6 (2021), 2378–2392.
\bibitem{b9} Akoglu L, Tong H, Koutra D. Graph based anomaly detection and description: a survey[J]. Data mining and knowledge discovery, 2015, 29: 626-688.
\bibitem{b10} Zhang K, Zhang C, Peng X. Putracead: Trace anomaly detection with partial labels based on GNN and Pu Learning[C]. 2022 IEEE 33rd International Symposium on Software Reliability Engineering (ISSRE). IEEE, 2022: 239-250.
\bibitem{b11} Tavares G M, Junior S B. Process mining encoding via meta-learning for an enhanced anomaly detection[C]. European Conference on Advances in Databases and Information Systems. Cham: Springer International Publishing, 2021: 157-168.
\bibitem{b12} Wang Q, Pang G, Salehi M. Cross-domain graph anomaly detection via anomaly-aware contrastive alignment[C]. Proceedings of the AAAI Conference on Artificial Intelligence. 2023, 37(4): 4676-4684.
\bibitem{b13} Kaize Ding, Qinghai Zhou, Hanghang Tong, and Huan Liu. 2021. Few-shot network anomaly detection via cross-network meta-learning. In Proceedings of the Web Conference 2021. 2448–2456.
\bibitem{b14} Daixin Wang, Jianbin Lin, Peng Cui, Quanhui Jia, Zhen Wang, Yanming Fang, Quan Yu, Jun Zhou, Shuang Yang, and Yuan Qi. 2019. A semi-supervised graph attentive network for financial fraud detection. In 2019 IEEE International Conference on Data Mining (ICDM). IEEE, 598–607.
\bibitem{b15} Tang J, Li J, Gao Z, Jia Li. Rethinking graph neural networks for anomaly detection[C]. International Conference on Machine Learning. PMLR, 2022: 21076-21089.
\bibitem{b16} Wang C, Pan S, Long G. Mgae: Marginalized graph autoencoder for graph clustering[C]. Proceedings of the 2017 ACM on Conference on Information and Knowledge Management. 2017: 889-898.
\bibitem{b17} Chen B, Zhang J, Zhang X. Graph Contrastive Learning for Anomaly Detection[J]. IEEE TKDE, 2021.
\bibitem{b18} Duan J, Wang S, Zhang P. Graph anomaly detection via multi-scale contrastive learning networks with augmented view[C]. Proceedings of the AAAI Conference on Artificial Intelligence. 2023, 37(6): 7459-7467.
\bibitem{b19} Ding K, Wang J, Caverlee J, H Liu. Meta propagation networks for graph few-shot semi-supervised learning[C]. Proceedings of the AAAI Conference on Artificial Intelligence. 2022, 36(6): 6524-6531.
\bibitem{b20} Zheng Y, Jin M, Liu Y, et al. From unsupervised to few-shot graph anomaly detection: A multi-scale contrastive learning approach[J]. IEEE TKDE, 2022.
\bibitem{b21} Y Zhu, Y Xu, F Yu, Q Liu, S Wu, L Wang. Graph contrastive learning with adaptive augmentation[C]. Proceedings of the Web Conference 2021. 2021: 2069-2080.
\bibitem{b22} T Zhao, Y Liu, L Neves, O Woodford, M Jiang. Data augmentation for graph neural networks[C]. Proceedings of the aaai conference on artificial intelligence. 2021, 35(12): 11015-11023.
\bibitem{b23} Frederickson G N, Ja’Ja’ J. Approximation algorithms for several graph augmentation problems[J]. SIAM Journal on Computing, 1981, 10(2): 270-283.
\bibitem{b24} Gallicchio C, Micheli A. Fast and deep graph neural networks[C]//Proceedings of the AAAI conference on artificial intelligence. 2020, 34(04): 3898-3905.
\bibitem{b25} D Bo, X Wang, C Shi, H Shen. Beyond low-frequency information in graph convolutional networks[C]. Proceedings of the AAAI Conference on Artificial Intelligence. 2021, 35(5): 3950-3957.
\bibitem{b26} Prithviraj Sen, Galileo Namata, Mustafa Bilgic, Lise Getoor, Brian Galligher, and Tina Eliassi-Rad. 2008. Collective classification in network data. AI magazine 29, 3 (2008), 93–93.
\bibitem{b27} Giles C L, Bollacker K D, Lawrence S. CiteSeer: An automatic citation indexing system[C]. Proceedings of the third ACM conference on Digital libraries. 1998: 89-98.
\bibitem{b28} Srijan Kumar, Xikun Zhang, and Jure Leskovec. 2019. Predicting dynamic embedding trajectory in temporal interaction networks. In Proceedings of the 25th ACM SIGKDD international conference on knowledge discovery \& data mining. 1269–1278.
\bibitem{b29} Will Hamilton, Zhitao Ying, and Jure Leskovec. 2017. Inductive representation learning on large graphs. In NeurIPS.
\bibitem{b30} Yingtong Dou, Zhiwei Liu, Li Sun, Yutong Deng, Hao Peng, and Philip S Yu. 2020. Enhancing graph neural network-based fraud detectors against camouflaged fraudsters. In Proceedings of the 29th ACM International Conference on Information \& Knowledge Management. 315–324.
\bibitem{b31} Veličković P, Cucurull G, Casanova A, et al. Graph attention networks[J]. ICLR, 2017.
\bibitem{b32} Huang J, Ling C X. Using AUC and accuracy in evaluating learning algorithms[J]. IEEE Transactions on knowledge and Data Engineering, 2005, 17(3): 299-310.



\end{thebibliography}
\end{document}